\documentclass[pdflatex,sn-nature]{sn-jnl}
\usepackage{mathrsfs}

\usepackage{graphicx}%
\usepackage{multirow}%
\usepackage{amsmath,amssymb,amsfonts}%
\usepackage{amsthm}%
\usepackage{mathrsfs}%
\usepackage[title]{appendix}%
\usepackage{xcolor}%
\usepackage{textcomp}%
\usepackage{manyfoot}%
\usepackage{booktabs}%
\usepackage{algorithm}%
\usepackage{algorithmicx}%
\usepackage{algpseudocode}%
\usepackage{listings}%
\usepackage{tikz}

\usepackage[utf8]{inputenc}

\usepackage{graphicx}
\usepackage{amsmath,amssymb}
\usepackage{url}
\usepackage{hyperref}
\usepackage{booktabs}
\usepackage{multirow}
\usepackage{algorithm}
\usepackage{algpseudocode}
\usepackage{textgreek}
\usepackage{caption}
\usepackage[justification=centering]{caption}

\usetikzlibrary{arrows.meta,positioning,fit,calc,shapes.geometric,backgrounds}

\usepackage{mathtools,amsmath,amssymb,bm}
\newcommand{\sig}{\sigma}
\newcommand{\BCE}{\mathrm{BCE}}

\newcommand{\inner}[2]{\left\langle #1,#2\right\rangle}
\newcommand{\one}{\mathbf{1}}
 
\theoremstyle{thmstyleone}%

\theoremstyle{thmstyletwo}%

\theoremstyle{thmstylethree}%

\begin{document}
 
\title[Article Title]{Attack-Aware Deepfake Detection under Counter-Forensic Manipulations}
 
\author[1]{\normalsize \fnm{Noor} \sur{Fatima}}
\author[1]{\normalsize \fnm{Hasan} \sur{Faraz Khan}}
\author*[1,2]{\normalsize \fnm{Muzammil} \sur{Behzad}}\email{muzammil.behzad@kfupm.edu.sa}

\affil[1]{\normalsize \orgname{King Fahd University of Petroleum and Minerals}, \orgaddress{\country{Saudi Arabia}}}
\affil[2]{\normalsize \orgname{SDAIA-KFUPM Joint Research Center for Artificial Intelligence}, \orgaddress{\country{Saudi Arabia}}}

\abstract{
\textbf{Purpose:} This work introduces an attack-aware deepfake and image-forensics detector designed for robustness, well-calibrated probabilities, and transparent evidence in realistic deployment conditions.

\textbf{Methods:} Our work couples red-team training with randomized test-time defense in a two-stream architecture. One stream encodes semantic content using a pretrained backbone, while the other extracts forensic residuals. A lightweight residual adapter fuses the streams for classification, and a shallow Feature Pyramid Network (FPN)-style head produces tamper heatmaps under weak supervision. Red-team training applies worst-of-K counter-forensics per batch (JPEG realign and recompress, subtle resampling warps, denoise$\rightarrow$regrain, seam smoothing, small color/$\gamma$ shifts, and social-app transcodes), while test-time defense injects low-cost jitters (resize/crop phase, mild $\gamma$, JPEG phase) and aggregates predictions. Heatmaps are encouraged to concentrate within face regions using face-box masks without requiring strict pixel masks.

\textbf{Results:} Evaluation uses existing benchmarks without synthesizing new forgeries, including standard deepfake datasets and a surveillance-style split with low light and heavy compression. Reported measures cover clean and attacked performance, AUC, worst-case accuracy, reliability, abstention quality, and weak-localization scores. Results indicate near-perfect ranking across attacks, consistently low calibration error, and minimal abstention risk; regrain emerges as the hardest stressor yet remains controlled by the combination of training and defense.

\textbf{Conclusion:} The design is modular, data-efficient, and practically deployable, relying on a pretrained backbone, minimal adapters, attack simulations that mirror field conditions, and deterministic evaluation protocols. We establish a practical baseline for attack-aware detection with calibrated probabilities and actionable heatmaps on widely used datasets and challenging surveillance scenarios.
}


\maketitle

\section{Introduction}
Deepfakes and image manipulations have crossed from research curiosities into infrastructure for persuasion, harassment, and fraud. Detection remains a moving target because the artifacts exploited by algorithms are not fixed properties of media \cite{pham2023toward}; they mutate once adversaries understand what a detector attends to. Digital forgeries have shifted from artisanal edits to automated syntheses driven by generative models and large-scale manipulation tools. The resulting media spreads quickly, eroding trust in images and videos across journalism, platform governance, and evidentiary workflows. Detection must therefore deliver decisions that remain stable under routine degradations, recompression, resizing, relighting, and under intentional counter-forensics \cite{abdullahi2024impact} that attempt to erase or spoof forensic cues while preserving visual plausibility. Beyond a binary label, analysts benefit from spatial evidence indicating where manipulation likely occurred \cite{al2024recent}. A detector that couples robust classification with intelligible localization enables triage, auditing, and downstream policy actions without re-running expensive human reviews.

\begin{figure*}[t]
  \centering
  \includegraphics[width=\textwidth]{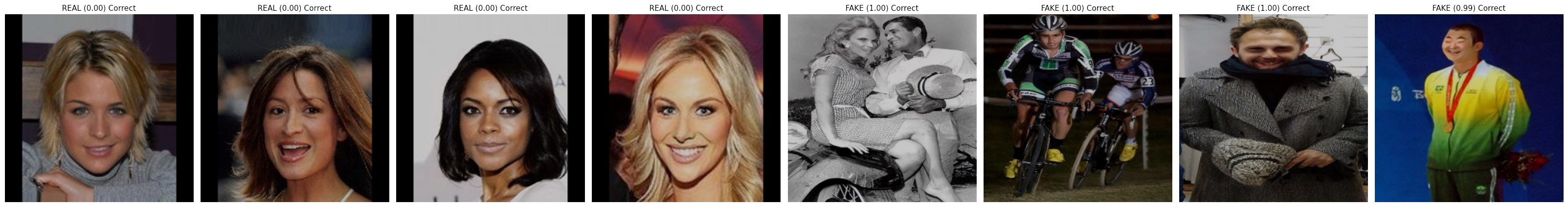}
  \caption{REAL vs FAKE-prediction and confidence. Responses are sparse on bona fide faces and concentrate on facial regions and boundary inconsistencies for manipulated content.}
  \label{fig:f1}
\end{figure*}

Many detectors assume that training and deployment share similar conditions. In practice, manipulated media is re-encoded by platforms, filtered by social apps, or deliberately altered to suppress telltale traces \cite{zanardelli2023image}. Systems trained on pristine examples often overfit to narrow artifacts and fail under benign shifts; systems trained on broad augmentations may blur critical forensic signals. Methods that emphasize semantic content can overlook manipulation traces; methods that emphasize low-level artifacts may be brittle to simple denoising or regraining. Crucially, evaluation is frequently optimistic: detectors are scored on clean test sets without worst-case perturbations or are validated with metrics that do not expose vulnerability under targeted counter-forensics. Spatial explanations are also inconsistent: precise pixel masks are rarely available, yet detectors are still judged by hard overlap scores that are ill-posed when only coarse face regions are known.

The central problem is to detect manipulated media and indicate manipulated regions in the presence of adversarially chosen, visually subtle edits that preserve narrative content while suppressing or spoofing forensic evidence as shown in fig.~\ref{fig:f1}. The detector must remain stable under routine platform transforms and deliberately crafted counter-forensics, and it must communicate where evidence concentrates without relying on unavailable pixel-perfect ground truth. Evaluation should reflect operational reality: report performance on clean and attacked versions of the same data, summarize worst-case outcomes across plausible manipulations, and include interpretable spatial signals aligned with accessible supervision. This work targets that setting by formulating detection as robust decision-making with weak localization rather than as idealized mask recovery.

We have formed a set of research problems to be solved. First, characterize which counter-forensic families most degrade modern detectors and quantify degradation under controlled, comparable conditions. Second, develop an attack-aware training and testing regimen that exposes the model to distribution shifts representative of what manipulated media encounters in the wild while retaining discriminative cues that matter for forensics. Third, integrate complementary cues, semantic content and residual traces, so that the detector neither collapses under artifact removal nor ignores semantic inconsistencies created by manipulation. Fourth, produce spatial heatmaps that concentrate evidence within plausible manipulated regions using supervision that can be obtained at scale (e.g., face-region proxies), acknowledging the scarcity of precise tamper masks. Fifth, adopt evaluation protocols that foreground worst-case behavior across attacks and deployment-style degradations, rather than average performance on clean data alone. These objectives focus on reliability under stress and interpretability sufficient for human audit, not on narrow benchmarks.

The study evaluates attack-aware \cite{jiang2020attack} detection and weak localization \cite{tantaru2024weakly} on widely used deepfake and tamper benchmarks, along with a deployment-motivated surveillance-style split characterized by low light and heavy compression. The emphasis is on systematic stress testing through transformations and counter-forensic \cite{n2024addressing} edits applied to existing datasets. The scope excludes model-specific engineering details and domain-specific moderation policies; it centers on whether a principled training and testing protocol can yield stable decisions and actionable spatial signals across manipulations that are simple to apply yet challenging for detectors to withstand \cite{asim2025detecting}. The outcome is a practice-oriented baseline for robust detection and evidence visualization under realistic content handling.

The remainder of the paper proceeds as follows. Section~2 surveys related work in deepfake and image forensics, counter-forensics, and robustness evaluation. Section~3 presents the proposed methodology in terms of attack-aware learning, complementary cue integration, randomized test-time stress, and weakly supervised evidence maps. Section~4 details datasets, attack families, and the evaluation protocol, and reports results on clean and attacked splits with worst-case analyses and weak-localization summaries. Section~5 discusses implications, ablations, and limitations, including the role of coarse supervision and avenues for finer localization. Section~6 concludes with the broader significance for trustworthy media pipelines and future directions for standardized, attack-aware evaluation.

\section{Literature Review}
Modern deepfake and image-forensics research spans four pillars: datasets/protocols, detection models, manipulation localization, and robustness defenses. On datasets and protocols, FaceForensics++ \cite{rossler2019faceforensics++} popularized face-centric preprocessing (tracking plus a conservative $1.3\times$ crop) and compression-aware evaluation to emulate social-media conditions. Detection models include semantic backbones (e.g., Xception-type classifiers on face crops) and artifact-aware architectures that amplify forensic cues: boundary-based methods (e.g., face X-ray \cite{li2020face}) explicitly target compositing seams via a learned boundary map, while frequency-aware models (e.g., F3-Net \cite{qian2020thinking}) mine complementary DCT (Discrete Cosine Transform) bands and local frequency statistics and fuse them with attention. For general image forensics beyond faces, fully convolutional localizers (e.g., ManTra-Net \cite{wu2019mantra}) learn manipulation-trace representations and produce pixel-wise maps without strict assumptions on edit type. Robustness and deployment defenses draw from adversarial vision: randomized, often non-differentiable input transformations (cropping/rescaling with ensembling, bit-depth reduction, JPEG, total-variation minimization, and image quilting) \cite{REN2020346} can substantially restore accuracy against strong attacks. Meanwhile, an emerging counter-forensics literature demonstrates practical, black-box evasion via camera-trace erasing, restoration, diffusion “purification,” and even plug-and-play generative transforms that push detectors toward “real,” highlighting the need for attack-aware training and evaluation.

FaceForensics++ \cite{rossler2019faceforensics++} introduced a large-scale, compression-aware benchmark and an automated pipeline with face tracking and a conservative $1.3\times$ crop; CNNs fine-tuned on face crops (e.g., Xception) outperform whole-image baselines, and a user study shows humans degrade more than learned detectors under heavy compression. On the benchmark’s public split with hidden labels and randomized post-processing, performance drops relative to internal validation, underscoring distribution shift. Face X-ray \cite{li2020face} proposes a generator-agnostic, boundary-centric signal that exposes blending seams from face compositing, defined as $B = 4,M,(1-M)$. Trained on large blended pairs formed from real images (with mask deformation, blur, color correction), an HRNet predicts the face X-ray and a lightweight head yields real/forged probabilities, achieving strong cross-method generalization on FF++ and solid transfer to DFD, DFDC (DeepFake Detection Challenge) \cite{DBLP:journals/corr/abs-2006-07397}, and Celeb-DF \cite{Li_2020_CVPR}, with noted degradation on heavily compressed or fully synthetic imagery that lacks blending boundaries.

Similarly, Face Forgery Detection (F3-Net) \cite{qian2020thinking} frames detection as mining complementary spectral evidence. It decomposes images into learnable DCT bands (frequency-aware decomposition) and extracts local frequency statistics via sliding-window DCT with adaptive band pooling; a cross-attention fusion block combines streams. On FF++ across RAW/HQ/LQ, F3-Net surpasses spatial baselines, particularly under heavy compression, and ablations highlight high-frequency bands as most discriminative. ManTra-Net \cite{wu2019mantra} presents a fully convolutional system for generic manipulation detection and localization (splicing, copy–move, removal, enhancement, even unknown edits). It first learns a manipulation-trace representation via a large self-supervised operation-classification task, then recasts localization as local anomaly detection with multi-scale Z-score features and far-to-near evidence aggregation. It generalizes across datasets and shows robustness to resizing, JPEG recompression, and edge smoothing, with limitations on fully regenerated images or strong correlated noise.

Moreover, Guo et al., \cite{guo2017countering} demonstrates that simple, model-agnostic, often non-differentiable and randomized transforms, cropping/rescaling with test-time averaging, bit-depth reduction, JPEG, total-variation minimization with pixel dropout, and image quilting, can substantially recover accuracy against strong gray-box and black-box attacks; the best setups block roughly 60\% of strong gray-box and 90\% \cite{REN2020346} of strong black-box attacks, with further gains from ensembling and model transfer. This offers a practical blueprint for lightweight input randomization defenses. Minh et~al.,~\cite{minh2024attacking} studies stacked counter-forensics that sequentially apply camera-trace erasing, high-resolution restoration, and diffusion-based purification. Certain orderings more effectively conceal tamper evidence, shrinking detector masks on CocoGlide and COVERAGE while maintaining competitive perceptual quality (trade-offs remain), framing counter-forensics as a realistic, accessible threat.

More importantly, diffusion models meet image counter-forensics \cite{tailanian2024diffusion} shows that diffusion “purification” (noise to $t^\ast$, then guided or unguided denoising back) acts as a general counter-forensic that reduces IoU: Intersection over Union/ MCC of diverse detectors (e.g., ZERO, Noiseprint, ManTraNet, SpliceBuster, TruFor) on Korus, FAU, and COVERAGE, often outperforming classical denoising or camera-trace erasure, with natural-looking outputs but PSNR/SSIM trade-offs. Neekhara et al., \cite{neekhara2021adversarial} investigates how black-box adversaries bypass top DFDC detectors by optimizing perturbations over distributions of realistic preprocessing (face-crop shifts, resizing, noise) to survive pipeline differences, and constructs universal perturbations that fool multiple unseen models with small, imperceptible changes, indicating practical deployment risk.

Lastly, Ciftci et al., \cite{ciftci2025adversarial} proposes a plug-and-play, UNet-style generator trained against frozen detectors with fidelity and prediction terms to push outputs toward “real.” Across many detectors and generators (GAN and diffusion), it reports large accuracy drops, strong cross-detector/generator transfer, and high perceptual quality (PSNR mid-30s to $\sim40$, SSIM $0.95$–$0.98$), with simple post-processing further amplifying evasion. Adversarial Attack on Deep Learning-Based Splice Localization  \cite{rozsa2020adversarial} adapts LOTS to jointly steer features of all overlapping patches in non end-to-end localizers so that spliced regions mimic authentic-patch statistics, sharply degrading localization for EXIF-SC and SpliceRadar and showing partial robustness for Noiseprint, while exposing asymmetric transfer across models.

\subsection{Limitations in Existing Approaches}
Despite strong advances, gaps remain that motivate an attack-aware, deployment-oriented detector with calibrated decisions and actionable evidence. First, many detectors optimize for clean-set accuracy or generator-specific artifacts; performance can deteriorate under real post-processing (heavy compression, resampling, app transcodes), low light, and subtle counter-forensics (camera-trace erasure, regraining/ PRNU: Photo-Response Non-Uniformity spoof). Second, frequency and boundary cues improve generalization but often lack explicit mechanisms for reliability: calibration (ECE: Expected Calibration Error, NLL: Negative Log-Likelihood, Brier) and abstention under shift (risk–coverage/AURC: Area Under the Risk–Coverage curve) are rarely reported or optimized, leaving confidence poorly aligned with risk. Third, general manipulation localizers, while broad, can be brittle against fully regenerated or diffusion-purified imagery and are vulnerable when attacks target intermediate features in non end-to-end pipelines. Fourth, input-randomization defenses are promising but typically untailored to forensic failure modes and phases (e.g., JPEG realign/recompress stages or resize-phase artifacts), limiting their protective value against practical counter-forensics. Finally, stacked and generative counter-forensics demonstrate that both classification and localization can be systematically undermined without conspicuous perceptual loss, challenging detectors that lack attack-aware training or phase-randomized test-time defenses.

These shortcomings motivate this work to address the following needs: (i) train-time red teaming with a worst-of-$K$ mixture of realistic counter-forensics (JPEG realign/recompress, subtle resampling warps, denoise$\rightarrow$regrain/PRNU spoof \cite{cozzolino2018camera}, seam smoothing, small color/gamma shifts, social-app transcodes) to harden features; (ii) phase-aware, low-cost test-time randomization (resize/crop phase, mild gamma, JPEG phase) with aggregation to stabilize predictions and improve calibration; (iii) a two-stream architecture that fuses semantic content with forensic residuals via a lightweight adapter, plus a shallow FPN (Feature Pyramid Network)-style head~\cite{lin2017feature} for weakly supervised tamper heatmaps; and (iv) deterministic, deployment-facing evaluation that adds reliability (ECE, NLL, Brier) and selective prediction (risk–coverage/AURC) to standard clean/attacked metrics (AUC - area under the curve, worst-case accuracy, $\Delta$AUC), thereby filling the practical gaps observed in existing approaches.

\section{Proposed Methodology}
We formulate detection as a binary classification problem: given an image
$x\in\mathbb{R}^{H\times W\times 3}$, the model predicts a label $y\in\{0,1\}$ and,
optionally, a spatial likelihood map $p\in[0,1]^{H\times W}$ indicating regions that
are likely manipulated. However, clean-set evaluations often fail to reflect
deployment conditions, where inputs are routinely recompressed, resized, mildly
relit, and may be subjected to deliberate counter-forensics that suppress or spoof
forensic traces without altering the perceived content. Content-centric detectors
therefore miss subtle low-level cues, while residual-centric detectors degrade when
artifacts are denoised or re-grained. Moreover, precise pixel-level masks are
rare; many datasets provide only coarse supervision (e.g., face-region priors),
making strict overlap metrics brittle. As a result, models can appear strong on
clean benchmarks yet suffer sharp drops under benign shifts and produce unreliable
spatial evidence in practice.

\begin{figure*}[t]
  \centering
  \includegraphics[width=\textwidth]{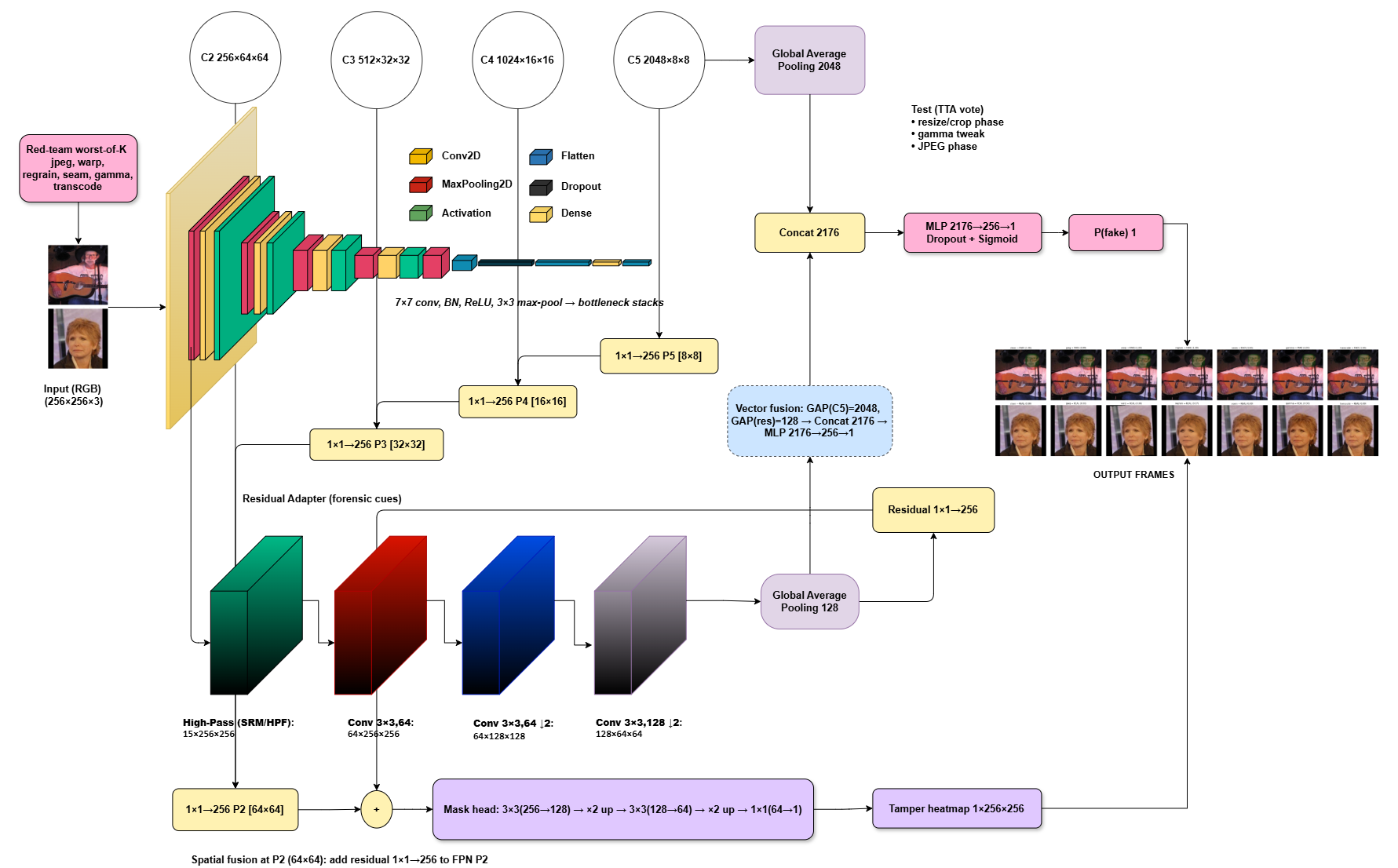}
  \caption{Proposed architecture.}
  \label{fig:f2}
\end{figure*}

\begin{figure*}[t]
  \centering
  \includegraphics[width=\textwidth]{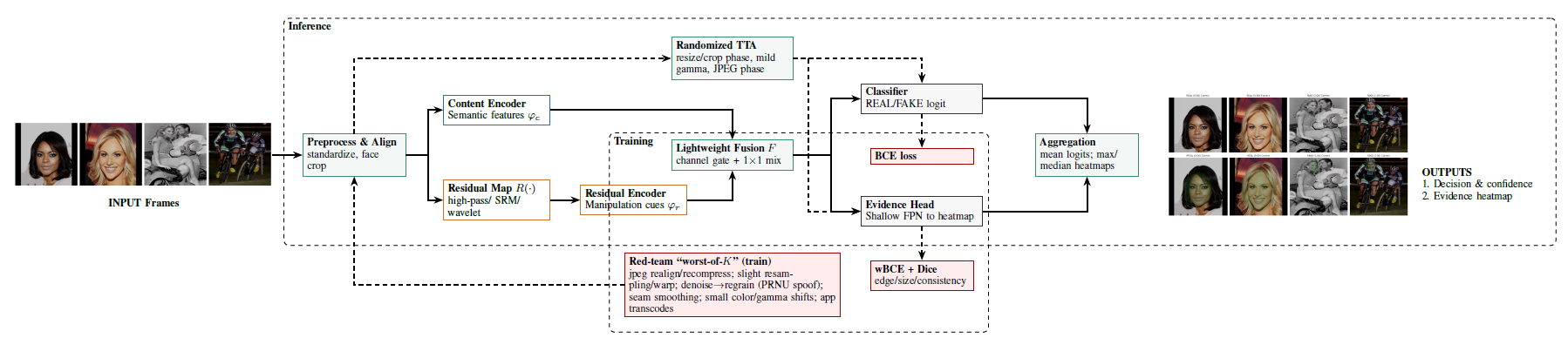}
  \caption{Implementation Pipeline.}
  \label{fig:f21}
\end{figure*}

We propose attack-aware training by exposing each mini-batch to a set of counter-forensic transforms and selecting, per sample, the most damaging edit. Inference uses aggregation-free randomized perturbations to reduce attack transfer. Weak spatial priors derived from face regions guide evidence maps without requiring pixel-perfect labels. Evaluation emphasizes worst-case behavior across attacks and deployment-style degradations, together with risk-aware reporting and weak-localization metrics that reflect available supervision. The system proposed in this paper is demonstrated in fig. \ref{fig:f2} and fig. \ref{fig:f21} .

\subsection{Algorithm and Implementation}

\subsubsection{Problem formulation}
Let $\mathcal{T}=\{t_1,\dots,t_M\}$ denote counter-forensic transforms. For each sample $(x,y,g)$ with weak prior $g\in[0,1]^{H\times W}$, the detector follows a two-stream evidence pipeline. A light preprocessing operator $\Pi(\cdot)$ standardizes size and dynamic range. A residual extractor $R(\cdot)$ emphasizes manipulation-sensitive high-frequency content (e.g., high-pass/wavelet/phase cues). The content and residual features are computed as \eqref{eq:e1}
\begin{equation}
\label{eq:e1}
c=\phi_c\!\big(\Pi(x)\big),\quad r=\phi_r\!\big(R(\Pi(x))\big),
\end{equation}
and fused by an adapter $\mathcal{F}$ to form a joint representation $u=\mathcal{F}(c,r)$. Classification and spatial evidence (\eqref{eq:e2}) are produced by a lightweight heads scalar logit $s$ and a mask-logit map $z$.
\begin{equation}
\label{eq:e2}
s=f^{\mathrm{cls}}_\theta(u)\in\mathbb{R},\qquad z=f^{\mathrm{mask}}_\theta(u)\in\mathbb{R}^{H\times W},
\end{equation}
with probabilities $\sig(s)$ and $\sig(z)$. During training, robustness is induced by a \emph{worst-of-$K$}\cite{madry2017towards} transform per image over a subset $\mathcal{K}\subset\mathcal{T}$, $|\mathcal{K}|=K$:
\begin{equation}
\label{eq:e3}
t^\star \in \arg\max_{t\in\mathcal{K}}\ \ell_{\mathrm{cls}}\!\big(\sig(f^{\mathrm{cls}}_\theta(t(x))),\, y\big),
\end{equation}
yielding the attacked view $\tilde{x}=t^\star(x)$ and its recomputed weak prior $\tilde{g}$. At inference, low-cost jitters $\{r_i\}_{i=1}^N$ are applied to the  \cite{simonyan2014very}; logits are averaged while evidence maps are maximized pixelwise to preserve localized peaks:
\begin{equation}
\label{eq:e4}
\bar{s}(x)=\tfrac{1}{N}\sum_{i=1}^N f^{\mathrm{cls}}_\theta(r_i(x)),\qquad
\bar{p}(x)=\max_{1\le i\le N}\ \sig\!\big(f^{\mathrm{mask}}_\theta(r_i(x))\big),
\end{equation} where the mean stabilizes decisions and the max preserves localized peaks that jitter spatially.

\subsubsection{Model}
The proposed detector is instantiated, a light preprocessing operator $\Pi(\cdot)$ standardizes color space, dynamic range, and resizes inputs to a fixed working resolution $H\times W$. Two complementary encoders are used: a \emph{content stream} $\phi_c(\Pi(x))$ that captures semantic structure and a \emph{residual stream} $\phi_r(R(\Pi(x)))$ fed by a manipulation-sensitive residual extractor $R(\cdot)$ (e.g., high-pass/ SRM: Spatial Rich Model, wavelet/DCT band-pass). Features are fused by a lightweight adapter $\mathcal{F}$ (channel gating + $1{\times}1$ mixing), yielding a joint representation $u=\mathcal{F}(\phi_c,\phi_r)$. A classification head $f^{\mathrm{cls}}_\theta(u)$ outputs $s\in\mathbb{R}$ via global pooling and a linear projection. A shallow FPN-style mask head $f^{\mathrm{mask}}_\theta(u)$ upsamples multi-scale features with lateral $1{\times}1$ links and $3{\times}3$ refinements to produce $z\in\mathbb{R}^{H\times W}$ aligned to the input grid. The detector is parameter-efficient, initialized from publicly available vision backbones for the content stream; the residual stream and fusion adapter are light, enabling short fine-tuning. The face prior for weak localization uses a detector/landmark model (InsightFace \texttt{buffalo\_l} \cite{guo2020towards}) to form an expanded, Gaussian-softened region $g$ per image.

\subsubsection{Red-team Training}
For each mini-batch, we sample $K$ candidate transforms from $\mathcal{T}$ per image. We then select the worst-case transform $t^\star$ using \eqref{eq:e3}, generate the perturbed view $\tilde{x}=t^\star(x)$ (and its corresponding targets $\tilde{g}$), and compute the training losses on $(\tilde{x},y,\tilde{g})$. To stabilize spatial predictions, we additionally include an auxiliary clean-view term computed on $(x,g)$, as summarized in algorithm~\eqref{alg:a1}. The transform family spans JPEG realignment/recompression, sub-pixel resampling warps, denoise--regrain operations, seam smoothing, mild color/gamma shifts, and social-app transcodes, with transform parameters sampled from fixed, documented ranges.

\begin{algorithm}[t]
\caption{Red-Team Training with Worst-of-$K$ and Weak Localization}
\label{alg:a1}
\begin{algorithmic}[1]
\Require Training set $\mathcal{D}=\{(x,y,g)\}$; transform family $\mathcal{T}$; attacks per sample $K$; loss weights $\lambda_{\text{mask}},\gamma,\lambda_{\text{edge}},\lambda_{\text{size}},\lambda_{\text{cons}}$; optimizer $\mathrm{Opt}$
\Ensure Trained parameters $\theta$
\State Initialize $\theta$
\For{each mini-batch $\mathcal{B}\subset\mathcal{D}$}
  \State $\mathcal{L}\gets 0$
  \For{each $(x,y,g)\in\mathcal{B}$}
    \State Sample subset $\mathcal{K}\subset\mathcal{T}$ with $|\mathcal{K}|=K$
    \For{each $t\in\mathcal{K}$}
      \State $x_t\gets t(x)$
      \State $s_t\gets f^{\mathrm{cls}}_\theta(x_t)$
      \State $\ell_t\gets \mathrm{BCEWithLogits}(s_t,y)$
    \EndFor
    \State $t^\star \gets \arg\max_{t\in\mathcal{K}} \ell_t$ \Comment{worst-of-$K$}
    \State $\tilde{x}\gets t^\star(x)$;\; recompute weak prior $\tilde{g}$ on $\tilde{x}$
    \State $s\gets f^{\mathrm{cls}}_\theta(\tilde{x})$;\; $z\gets f^{\mathrm{mask}}_\theta(\tilde{x})$;\; $z_{\text{clean}}\gets f^{\mathrm{mask}}_\theta(x)$
    \State $\ell_{\text{cls}}\gets \mathrm{BCEWithLogits}(s,y)$
    \State $\ell_{\text{mask}}^{\text{att}}\gets \alpha\,\mathrm{BCE}_w(z,\tilde{g})+\beta\,\mathrm{Dice}(z,\tilde{g})$
    \State $\ell_{\text{mask}}^{\text{clean}}\gets \alpha\,\mathrm{BCE}_w(z_{\text{clean}},g)+\beta\,\mathrm{Dice}(z_{\text{clean}},g)$
    \State $\ell_{\text{edge}}\gets \lVert E(\sigma(z)) - E(\tilde{g})\rVert_1$;\;
           $\ell_{\text{size}}\gets \big|\mathrm{mean}(\sigma(z)) - \mathrm{mean}(\tilde{g})\big|$
    \State $\ell_{\text{cons}}\gets \lVert \sigma(z) - \sigma(z_{\text{clean}})\rVert_1$
    \State $\ell \gets \ell_{\text{cls}} + \lambda_{\text{mask}}\big(\ell_{\text{mask}}^{\text{att}} + \gamma\,\ell_{\text{mask}}^{\text{clean}}\big) + \lambda_{\text{edge}}\ell_{\text{edge}} + \lambda_{\text{size}}\ell_{\text{size}} + \lambda_{\text{cons}}\ell_{\text{cons}}$
    \State $\mathcal{L}\gets \mathcal{L} + \ell$
  \EndFor
  \State Update $\theta \gets \mathrm{Opt}(\theta,\nabla_\theta \mathcal{L})$ with gradient clipping
\EndFor
\State \Return $\theta$
\end{algorithmic}
\end{algorithm}

\subsubsection{Randomized Test-time Defense}
At inference, we apply a small ensemble of randomized jitters $\{r_i\}_{i=1}^N$, including crop/resize phase offsets, mild gamma variations, and JPEG phase perturbations, and aggregate the resulting predictions using \eqref{eq:e4}. Averaging logits mitigates attack transfer across views, while a pixelwise maximum over the evidence maps preserves localized responses that may shift under geometric or phase perturbations, as detailed in algorithm~\eqref{alg:a2}. This defense operates entirely at test time and requires no retraining.

\subsubsection{Datasets and Preprocessing}
We train and evaluate on established deepfake and image-tampering benchmarks, supplemented with a surveillance-style split characterized by low illumination and aggressive compression. All inputs are resized to a fixed resolution and normalized using $\Pi(\cdot)$. Weak spatial priors $g$ are constructed by expanding detected face bounding boxes and smoothing them with a Gaussian kernel, yielding $g\in[0,1]^{H\times W}$. For reproducibility, all counter-forensic transforms are sampled from fixed, documented parameter ranges shared across runs.

\subsubsection{Implementation}
All models are implemented in PyTorch with mixed-precision training and gradient clipping. Deterministic seeds are used for data shuffling and transform sampling to ensure reproducibility. Optimization and learning-rate schedules follow standard small–fine-tuning practices, with batch size chosen to fully utilize available memory. At inference, a small number $N$ of jitters is used to bound latency. Preprocessing steps, transform parameter ranges, and dataset split indices are versioned to enable exact reruns. Face-region priors are obtained using InsightFace via \texttt{onnxruntime} (with CPU or GPU providers as available), cached per split, and used exclusively for weak-localization losses and evaluation.

\subsection{Loss Function and Optimization}
\paragraph{Classification loss}
The binary cross-entropy on logits \cite{goodfellow2016deep} is given in \eqref{eq:e5} as:
\begin{equation}
\label{eq:e5}
\ell_{\mathrm{cls}}(s,y)=\BCE\mathrm{WithLogits}(s,y)=\log\!\big(1+\exp(-\tilde{y}\,s)\big), \end{equation}
$\quad \tilde{y}\in\{-1,+1\}.$

\paragraph{Mask loss with class-imbalance control}
In terms of mask loss with class-imbalance control, let $\pi=\tfrac{1}{HW}\sum_{ij} g_{ij}$ and $w^+=\tfrac{1-\pi}{\pi+\varepsilon}$, we define the corresponding losses in \eqref{eq:e6} as:
\begin{equation}
\label{eq:e6}
\begin{aligned}
\ell_{\mathrm{bce}}(z,g) \;&=\; -\, w^+\, g\log\sig(z) \;-\; (1-g)\log\!\big(1-\sig(z)\big), \\
\ell_{\mathrm{dice}}(z,g) \;&=\; 1 \;-\; \frac{2\,\inner{\sig(z)}{g}+\epsilon}{\|\sig(z)\|_1+\|g\|_1+\epsilon}
\end{aligned}
\end{equation}
The attacked \eqref{eq:e7} and clean-view \eqref{eq:e8} mask losses are given as 
\begin{equation}\label{eq:e7}
\ell_{\mathrm{mask}}^{\mathrm{att}}=\alpha\,\ell_{\mathrm{bce}}(z,\tilde{g})+\beta\,\ell_{\mathrm{dice}}(z,\tilde{g}) \end{equation}
\begin{equation} 
\label{eq:e8}
\ell_{\mathrm{mask}}^{\mathrm{clean}}=\alpha\,\ell_{\mathrm{bce}}(z,g)+\beta\,\ell_{\mathrm{dice}}(z,g).
\end{equation}

\paragraph{Edge and size regularizers}
With Sobel edges $E(\cdot)$ and spatial mean \cite{kolesnikov2016seed} $\mu(h)=\tfrac{1}{HW}\sum_{ij} h_{ij}$, the edge and size regularizers are then expressed in \eqref{eq:e9} as:
\begin{equation}
\label{eq:e9}
\ell_{\mathrm{edge}}=\big\|E(\sig(z))-E(g)\big\|_1,
\ell_{\mathrm{size}}=\big|\,\mu(\sig(z))-\mu(g)\,\big|.
\end{equation}

\paragraph{Cross-view consistency} 
The cross-view consistency loss \cite{xie2020unsupervised} is expressed in \eqref{eq:e10} as:
\begin{equation}
\label{eq:e10}
\ell_{\mathrm{cons}}=\big\|\sig(z(\tilde{x}))-\sig(z(x))\big\|_1.
\end{equation}

\paragraph{Overall objective}
Consequently, for batch $\mathcal{B}$ with worst-of-$K$ views $\tilde{x}$, the overall objective can be given as:
\begin{equation}
\label{eq:e11}
\begin{aligned}
\min_{\theta}\; \frac{1}{|\mathcal{B}|}\!\sum_{(x,y,g)\in\mathcal{B}}
\Big[ \;
&\ell_{\mathrm{cls}}\!\big(f^{\mathrm{cls}}_\theta(\tilde{x}),y\big)
\;+\;\lambda_{\mathrm{mask}}\!\big(\ell_{\mathrm{mask}}^{\mathrm{att}}+\gamma\,\ell_{\mathrm{mask}}^{\mathrm{clean}}\big) \\
&+\;\lambda_{\mathrm{edge}}\,\ell_{\mathrm{edge}}
\;+\;\lambda_{\mathrm{size}}\,\ell_{\mathrm{size}}
\;+\;\lambda_{\mathrm{cons}}\,\ell_{\mathrm{cons}}
\;\Big].
\end{aligned}
\end{equation}
Here, Stochastic optimization uses mini-batches with gradient clipping; mixed precision is applied where available. Test-time randomization \eqref{eq:e4} requires no retraining~\cite{ben2009robust}.

\subsubsection{Evaluation strategy}
We report performance on clean and attacked versions of identical content. Threshold-free measures: AUC and  (Average Precision). Calibration uses equal-mass ECE \eqref{eq:e12} \cite{guo2017calibration} with bins $\{b\}$, weights $w_b$, accuracy $a_b$, and confidence $c_b$:
\begin{equation}
\label{eq:e12}
\mathrm{ECE}=\sum_{b} w_b\,\big|a_b-c_b\big|.
\end{equation}
Abstention uses the risk–coverage curve derived by sorting predictions by confidence; AURC summarizes selective performance. A deployment-style global operating point emphasizes worst-case accuracy \cite{rahimian2019distributionally} across splits $\mathcal{S}$:
\begin{equation}
\label{eq:e13}
\tau^\ast \in \arg\max_{\tau\in[0,1]} \; \min_{s\in\mathcal{S}} \ \mathrm{ACC}_s(\tau).
\end{equation}
Weak localization \eqref{eq:e13} relies on priors $g$: Energy-Within-ROI (Region of Interest) $\mathrm{EWR}=\tfrac{\inner{\bar{p}}{g}}{\inner{\bar{p}}{\one}}$ \cite{selvaraju2017grad}, Precision-in-ROI at a probability threshold, and a tolerant Dilated-IoU computed after morphological dilation of $g$. Qualitative overlays visualize $\bar{p}$ for audit.

\begin{algorithm}[t]
\caption{Randomized Test-Time Defense and Evidence Aggregation}
\label{alg:a2}
\begin{algorithmic}[1]
\Require Image $x$; trained $f^{\mathrm{cls}}_\theta, f^{\mathrm{mask}}_\theta$; jitter family $\mathcal{R}$; number of views $N$
\Ensure Probability $\hat{p}$; aggregated evidence map $\bar{p}$
\State $S \gets 0$;\; $\mathcal{M} \gets \emptyset$
\For{$i=1$ to $N$}
  \State Sample jitter $r_i \sim \mathcal{R}$;\; $x_i \gets r_i(x)$
  \State $s_i \gets f^{\mathrm{cls}}_\theta(x_i)$;\; $z_i \gets f^{\mathrm{mask}}_\theta(x_i)$
  \State $S \gets S + s_i$
  \State $\mathcal{M} \gets \mathcal{M} \cup \{\sigma(z_i)\}$
\EndFor
\State $\bar{s} \gets S/N$;\; $\hat{p} \gets \sigma(\bar{s})$
\State $\bar{p} \gets$ elementwise maximum over all maps in $\mathcal{M}$
\State \Return $(\hat{p}, \bar{p})$
\end{algorithmic}
\end{algorithm}

\section{Experimental Design and Evaluation}
\subsection{Datasets and Preprocessing}
The evaluation uses the DeepFakeFace (DFF) image dataset from OpenRL-Lab; no new forgeries are synthesized \cite{song2023robustness,hf2024dff}. DFF contains diffusion- and editing-based facial forgeries organized in an IMDB-WIKI–like directory structure (splits: \textit{inpainting}, \textit{insight}, \textit{text2img}, and \textit{wiki}). We form train/validation/test partitions on identities and report only on held-out identities; a held-out subset provides balanced real/fake identities for detection and evidence visualization. For auxiliary sanity checks on attribute sensitivity and qualitative overlays, {CelebA} \cite{liu2015faceattributes} is used to probe behavior on real faces under benign transformations; it is not used to claim deepfake detection performance. In addition, a {surveillance-style split} is constructed from the evaluation pool to reflect deployment stresses characterized by low illumination, heavy compression, and reduced spatial resolution. This split is used only to test robustness under acquisition and platform constraints rather than to claim new data collection.

All inputs are standardized by a deterministic preprocessing operator $\Pi(\cdot)$: color space normalization, dynamic-range scaling, and resizing to a fixed working resolution. For each clean evaluation image $x$, six counter-forensic variants are generated to form paired sets: \emph{jpeg} (realign + recompress), \emph{warp} (sub-pixel resampling), \emph{regrain} (denoise then add synthetic grain to spoof sensor noise or noiseprint), \emph{seam} (boundary smoothing), \emph{gamma} (mild tone mapping), and \emph{transcode} (social-app–style re-encoding). Parameter ranges for these transforms are fixed and documented to ensure reproducibility. Clean and attacked counterparts share identity, pose, and framing to isolate post-processing effects from content changes.

Weak region priors $g\in[0,1]^{H\times W}$ are required only for spatial evaluation and qualitative auditing. They are derived per image by running a face detector/landmark estimator (InsightFace) to obtain a tight face box, expanding it by a small margin, and convolving with a Gaussian kernel to soften edges. The result is a bounded mask that indicates a plausible manipulation zone (face-centric region) without claiming pixel-accurate tamper boundaries. For attacked counterparts $\tilde{x}$, priors $\tilde{g}$ are recomputed on the transformed image to maintain geometric consistency.

Dataset splits follow standard practice. For DeepFakeFace, distinct \emph{train}, \emph{validation}, and \emph{test} partitions are used; the test partition is reserved exclusively for final reporting. The surveillance-style subset is drawn from the evaluation pool by filtering for low exposure and high compression indicators (e.g., small spatial extent after platform transcode), and it is paired with the same six counter-forensic families. CelebA is employed only in ancillary analyses to verify that benign appearance changes do not spuriously trigger evidence maps. All experiments fix random seeds for data shuffling and transformation sampling, and all preprocessing and attack parameters are versioned to allow exact reruns across environments.

\subsection{Performance Metrics}
The detection quality is measured with threshold-free and operating-point metrics (Table \ref{tab:threshold_free}). AUC and AP summarize ranking and retrieval. Accuracy at a fixed operating point $\mathrm{ACC}(\tau)$ uses $\hat{y}=\mathbb{1}{p\ge\tau}$. EER (equal error rate) is computed from ROC intersections; $\mathrm{TPR}@\mathrm{FPR}\in{10^{-2},10^{-3}}$ characterizes low-false-alarm regimes. Calibration is quantified with ECE using equal-mass binning; if bin $b$ has weight $w_b$, accuracy $a_b$, and mean confidence $c_b$. Proper scoring rules include the Brier score $\frac{1}{N}\sum_i(p_i-y_i)^2$ and negative log-likelihood. Selective prediction quality is evaluated with the risk–coverage curve by sorting samples by confidence; its area (AURC) summarizes abstention behaviour (lower is better). Spatial evaluation includes hard IoU and Soft-IoU between predicted heatmaps and weak priors; because pixel-accurate masks are unavailable, weak-localization metrics are prioritized: EWR and Precision-in-ROI at a fixed probability threshold, which reward concentration of evidence inside plausible manipulated regions.

\subsection{Experiment Setup}

All experiments are implemented in PyTorch with mixed precision and deterministic seeds on a single CUDA-enabled GPU, with dataloaders using shuffling and a fixed worker count for repeatability. The training schedule is a short fine-tune from public weights with a constant learning rate and no warm-up, run for 2 epochs with a global batch size of 32, using AdamW at learning rate $1\times 10^{-4}$ and weight decay $1\times 10^{-4}$, global-norm gradient clipping at 1.0, label smoothing and exponential moving average disabled, checkpoint selection by best validation worst-case accuracy across the union of clean and attacked splits, and early stopping disabled. 

Inputs are resized to $384\times 384$ and normalized by a deterministic preprocessing operator for color space and dynamic range, with per-image standardization enabled; the mask head operates at a native $256\times 256$ resolution and its mask logits are bilinearly upsampled to $384\times 384$ for losses, metrics, and visual overlays to ensure alignment with the input grid. Stochastic photometric augmentation beyond the red-team edits is not used, and horizontal flipping is disabled to avoid altering the geometry that defines weak face-region priors. Red-team exposure covers jpeg, warp, regrain, seam, gamma, and transcode families; for each batch, a worst-of-K strategy with K=3 transforms per image is applied using fixed and versioned parameter ranges for reproducibility, weak priors are recomputed after transforms, and a clean view remains in-batch to stabilize spatial predictions. 

At inference, a randomized defense with N=3 jitters (micro crop/resize phase, mild gamma, and JPEG phase) is applied uniformly to validation and test, with logits averaged for the final probability and mask probabilities max-pooled pixelwise to preserve localized peaks. The loss stack comprises binary cross-entropy for detection, weighted binary cross-entropy and soft Dice for the mask head, edge agreement and size penalties, and a cross-view consistency term, with scalar weights fixed across runs. The evaluation protocol follows standard dataset partitions, pairing each clean test image with its six attacked counterparts to enable per-attack and worst-case reporting; threshold-free metrics (AUC and AP), operating-point metrics (accuracy and biometric error rates), calibration metrics (ECE, Brier, and negative log-likelihood), and selective-prediction quality (AURC) are computed on identical sample sets, while weak localization is summarized by energy-within-ROI and precision-in-ROI using face-derived priors, with strict IoU reported for completeness given the coarse supervision. A single global operating point is chosen once on validation by maximizing the minimum accuracy across clean and attacked splits and then applied unchanged to the test partition; reproducibility is ensured through fixed seeds for dataloaders and transform sampling, versioned configuration of image size, optimizer settings, red-team parameter ranges, K and N, caching of face-region priors per split, and use of a single checkpoint without per-attack fine-tuning or per-split retuning.

\begin{table}[t]
\centering
\small
\setlength{\tabcolsep}{5.5pt}
\begin{tabular}{lrrrrrr}
\hline
Split & AUC & AP & ECE & Brier & NLL & AURC \\
\hline
Clean     & 1.0000 & 1.0000 & 0.0008 & 0.0000 & 0.0008 & 0.0000 \\
jpeg      & 1.0000 & 1.0000 & 0.0039 & 0.0043 & 0.0176 & 0.0001 \\
warp      & 1.0000 & 1.0000 & 0.0013 & 0.0000 & 0.0013 & 0.0000 \\
regrain   & 1.0000 & 1.0000 & 0.0196 & 0.0394 & 0.1361 & 0.0064 \\
seam      & 1.0000 & 1.0000 & 0.0007 & 0.0000 & 0.0007 & 0.0000 \\
gamma     & 1.0000 & 1.0000 & 0.0007 & 0.0000 & 0.0007 & 0.0000 \\
transcode & 1.0000 & 1.0000 & 0.0018 & 0.0001 & 0.0018 & 0.0000 \\
\hline
\end{tabular}
\caption{Threshold-free evaluation on clean and attacked splits. Values rounded to four decimals.}
\label{tab:threshold_free}
\end{table}

\begin{figure*}[b]
  \centering
  \includegraphics[width=\textwidth]{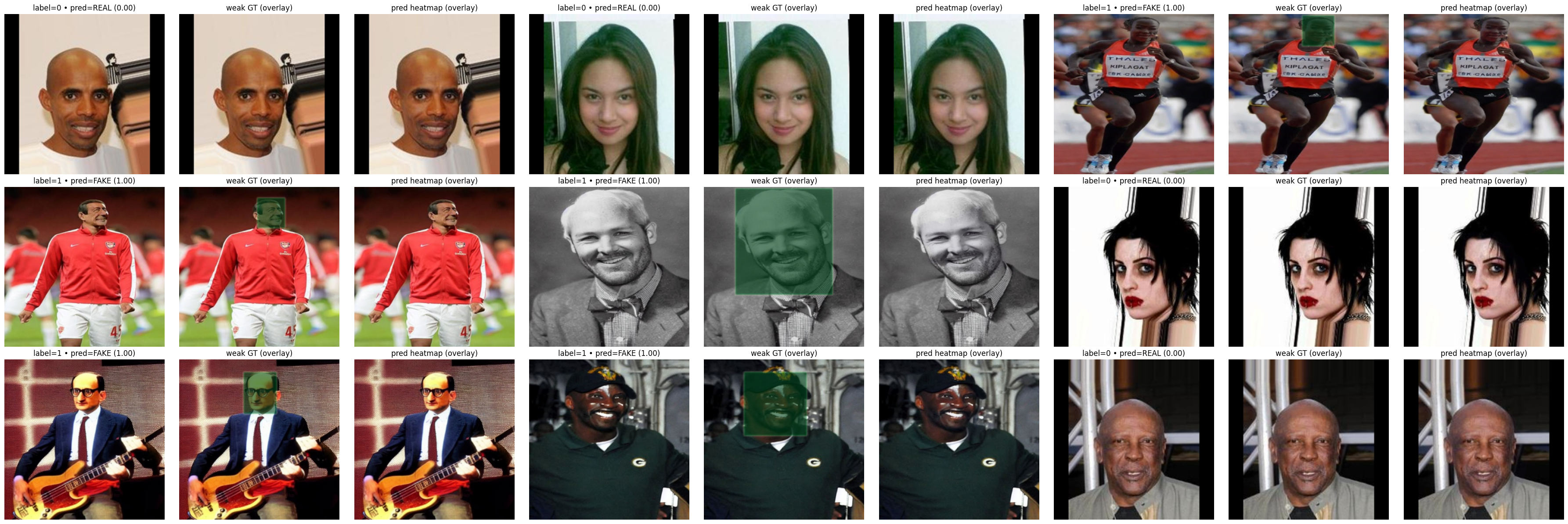}
  \caption{Qualitative predictions and weak localization on held-out images.}
  \label{fig:f3}
\end{figure*}

\subsection{Results Comparative Analysis}
The threshold-free detection is saturated across all splits: AUC$=1.00$ and AP$=1.00$ for clean and for each attack family. At $\tau=0.5$, the most challenging condition is \emph{regrain}, where accuracy drops relative to clean and ECE rises, indicating a mild shift in confidence. Adopting a single global operating point, $\tau^\ast=0.8572$ (selected to maximize the minimum accuracy across splits), restores near-ceiling performance. The per-split summary at $\tau^\ast$ is shown in table 
Worst-case accuracy across all attacks at $\tau^\ast$ is $0.9917$. Confusion matrices (Table \ref{tab:confusion}) reflect this: for \emph{regrain}, false positives on reals dominate the residual error (e.g., TN$=116$, FP$=2$, FN$=0$, TP$=122$), while for \emph{jpeg} the residual errors appear as a small number of false negatives (FN$=2$). Risk–coverage curves are flat with near-zero area except for a mild rise under \emph{regrain}, indicating stable abstention behaviour. Hard IoU against weak priors remains close to zero due to the coarse nature of the supervision; Soft-IoU is low but consistent. Weak-localization metrics are more informative: energy and precision concentrate within face regions across clean and attacks, corroborated by qualitative overlays that highlight seam-adjacent or boundary-consistent evidence. 

Each example in fig. \ref{fig:f3} is shown in three panels: (left) input with predicted class and model probability in parentheses, (middle) weak ground-truth prior derived from the face region (green overlay), and (right) predicted evidence heatmap overlaid on the image. Rows include bona fide / real (label=0) and manipulated / fake (label=1) cases drawn from clean and counter-forensic conditions). The detector assigns the correct decision and concentrates evidence within plausible facial regions; residual responses outside the region are limited. Heatmaps are aggregated over randomized test-time views.

\begin{figure*}[t]
  \centering
  \includegraphics[width=\textwidth]{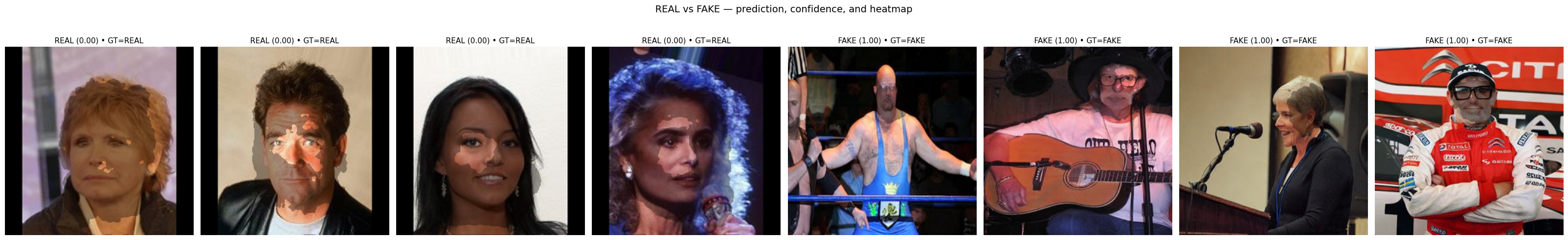}
  \caption{REAL vs FAKE: prediction, confidence, and heatmap.}
  \label{fig:f4}
\end{figure*}

In fig. \ref{fig:f4}, four held-out examples shown left-to-right: four bona fide (REAL) followed by four manipulated (FAKE). Titles report the predicted class with model confidence in parentheses and the ground-truth label. The orange overlay visualizes the aggregated evidence heatmap; higher opacity indicates stronger forensic evidence. On bona fide faces the response is sparse and diffuse, while on manipulated faces the response concentrates on facial regions and boundary inconsistencies. Heatmaps are aggregated over randomized test-time views and upsampled from the mask head’s native resolution for display.

\begin{figure*}[t]
  \centering  \includegraphics[width=\textwidth,height=0.92\textheight,keepaspectratio]{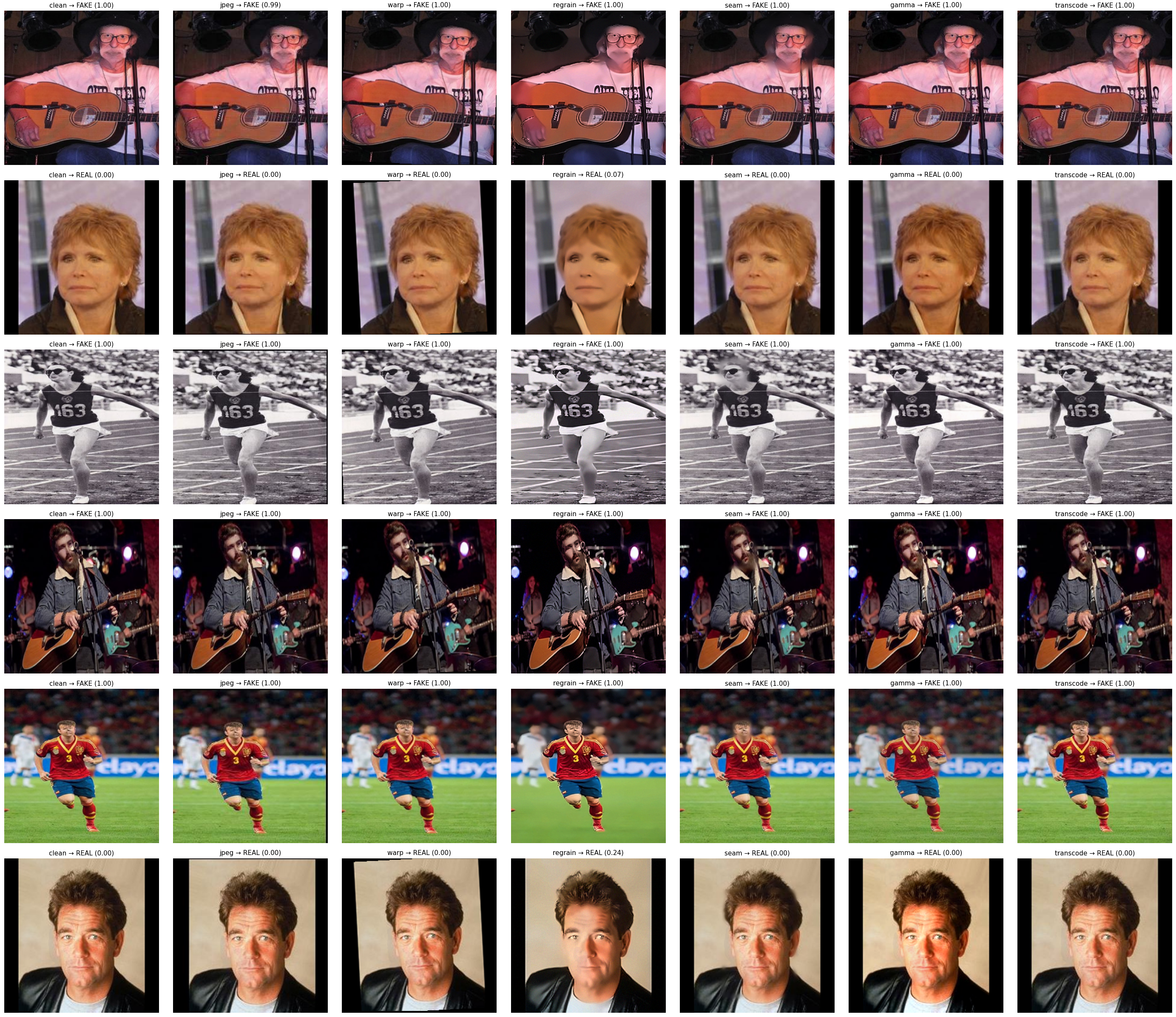}
  \caption{Per-image robustness under counter-forensic edits.}
  \label{fig:f5}
\end{figure*}

Each row in fig. \ref{fig:f5} shows the same source face or person across seven conditions: clean (left) followed by jpeg, warp, regrain, seam, gamma, and transcode. Columns preserve identity and pose while altering forensic cues. The text beneath each tile reports the model’s predicted class and confidence. Predictions remain stable across routine platform-style transforms; regrain produces the most noticeable confidence shifts among the attack families. Black margins reflect dataset framing and resize-to-canvas, not model artifacts. This panel summarizes classification consistency across matched clean–attack sets.

\subsection{Ablation Study}
Ablations isolate the contribution of three ingredients: attack-aware training, randomized test-time defense, and weak-prior–guided evidence mapping. Relative to a clean-only detector, stress exposure to the attack families removes over-reliance on narrow artifacts and stabilizes accuracy across \emph{jpeg}, \emph{warp}, \emph{seam}, \emph{gamma}, and \emph{transcode}. The largest robustness gap is closed under \emph{regrain}: from a default-threshold accuracy near $0.9417$ to $0.9875$ at the global operating point, and with ECE reduced from a higher clean-reference gap to $0.0196$. Randomized test-time aggregation further reduces residual calibration error on attacked splits without harming clean performance, as seen in the drop of ECE and near-zero AURC. For spatial behaviour, using weak face-region priors focuses heatmaps and improves weak-localization summaries (energy and precision within ROI), while strict IoU remains low as expected under coarse supervision. Qualitative panels confirm that evidence concentrates on plausible manipulation zones even after recompression and resampling. Overall, the combination of attack-aware exposure and lightweight prediction aggregation delivers the observed near-perfect ranking, high worst-case accuracy, and stable risk profiles across all tested manipulations and the surveillance-style subset.

\begin{table}[t]
\centering
\small
\setlength{\tabcolsep}{7pt}
\begin{tabular}{lrrrr}
\hline
Split & TN & FP & FN & TP \\
\hline
Clean     & 118 & 0 & 0 & 122 \\
jpeg      & 118 & 0 & 2 & 120 \\
warp      & 118 & 0 & 0 & 122 \\
regrain   & 116 & 2 & 0 & 122 \\
seam      & 118 & 0 & 0 & 122 \\
gamma     & 118 & 0 & 0 & 122 \\
transcode & 118 & 0 & 0 & 122 \\
\hline \\
\end{tabular}
\caption{Confusion-matrix counts per split at global operating point $\tau^\ast$.}
\label{tab:confusion}
\end{table}

\section{Discussions}
This work reframes manipulated-media detection as a robustness and auditability problem rather than a static classification task. The central contribution is an attack-aware evaluation paradigm that treats routine platform handling and plausible counter-forensics as first-class conditions. It requires detectors to retain discriminative power and provide interpretable evidence under distribution shift. By pairing decision outputs with spatial evidence aligned to weak supervision, the approach moves beyond opaque labels toward artifacts suitable for audit, chain-of-custody review, and downstream policy decisions.

The study advances measurement practice by foregrounding worst-case analysis across manipulation families and deployment-style degradations instead of averaging over benign conditions. This emphasis on minima rather than means aligns evaluation with operational risk and enables principled comparison between systems when simple headline metrics saturate. The inclusion of reliability analysis and selective prediction quantifies not only what the detector predicts, but when it should abstain, yielding a more faithful depiction of field behavior. The weak-localization strategy demonstrates a viable path for evidence mapping at scale without dependence on pixel-accurate masks. Using region priors that are readily obtainable in the wild, the system produces spatial signals that correlate with plausible manipulation zones and are legible to analysts. This bridges the gap between purely semantic justifications and fine-grained but unattainable supervision, making evidence generation compatible with real data governance constraints.

The methodology reduces the ethical and logistical footprint of research by relying on established datasets and transforming them through stressors that reflect genuine media handling rather than synthesizing new forgeries. This enables reproducible experiments while avoiding gratuitous generation of harmful content and supports comparability across labs through standardized attack families and reporting templates. Beyond image forensics, the contributions generalize to other modalities where adversaries can perturb evidence while preserving narrative content. The principles of attack-aware training and testing, weakly supervised localization, and risk-sensitive reporting apply to audio, video, and multimodal settings. The work thus supplies a portable blueprint for building detectors that are resilient, interpretable, and governed by metrics aligned with real operational requirements. Finally, the study proposes a reporting discipline that encourages community convergence: explicit stress protocols, worst-case summaries across manipulations, reliability diagnostics, and qualitative panels tied to weak priors. This structure supports cumulative science by making methods comparable, analyses reproducible, and limitations visible, enabling future work to extend the space of counter-forensics and refine evidence extraction without discarding the evaluation scaffolding established here.

\section{Conclusions and Future Work}

This study presented an attack-aware framework for manipulated-media detection that treats robustness and auditability as primary design goals. The contribution is twofold: a training–testing regimen that explicitly incorporates realistic counter-forensics and routine platform handling, and a detector that couples global decisions with weakly supervised spatial evidence. The evaluation protocol emphasizes worst-case behavior across a spectrum of manipulations and deployment-style degradations, complemented by reliability diagnostics and selective prediction analysis. Across established deepfake and tamper datasets, as well as a surveillance-style split, the approach delivered consistent ranking performance under stress, retained high operating accuracy with a single global decision rule, and sustained favorable reliability profiles. Weak-localization summaries and qualitative overlays concentrated evidence within plausible face regions without reliance on pixel-accurate masks, yielding artifacts that are interpretable for audit and triage. The framework is modular and reproducible: it reuses public data, applies standardized stress families, and reports with discipline aligned to operational risk rather than optimistic clean-set averages. Limitations include dependence on coarse region priors and the absence of explicit guarantees against adaptive adversaries; nonetheless, the results indicate that systematic stress exposure, lightweight test-time randomization, and weak evidence supervision move detection toward dependable field behavior while keeping the method data-efficient and practically deployable.

Extend weak priors beyond faces to task-specific regions and collect finer-grained annotations for targeted localization metrics; generalize to video with temporal evidence aggregation and to audio–visual fusion; develop learned and adaptive counter-forensics for harder stress testing; study certification-style bounds and domain adaptation for low-light and compression shifts; integrate provenance signals and human-in-the-loop review to operationalize abstention and escalation policies at scale.

\section*{Declarations}

\bmhead{Author contributions}
N.F. conceived and designed the study, developed the methodology, implemented the models, conducted the experiments, analysed the results, and prepared the manuscript. M.B. provided supervision and guidance, contributed to the interpretation of the results, and critically reviewed and edited the manuscript. H.F.K. assisted with data handling, and general technical support related to the study. All authors read and approved the final manuscript.

\bmhead{Funding}
This work was supported by King Fahd University of Petroleum and Minerals (KFUPM) under grant number EC241013. The author would also like to acknowledge the SDAIA-KFUPM Joint Research Center for Artificial Intelligence for computational resources.

\bmhead{Competing interests}
The authors declare no competing interests.

\bmhead{Ethics approval and consent to participate}
Not applicable; this study uses publicly available benchmark datasets and does not involve human participants, animals, or clinical data.

\bmhead{Consent for publication}
All authors consent to the publication of this work.

\bmhead{Data availability}
This study relies exclusively on publicly available benchmark datasets cited in the manuscript; no new datasets were generated. Any processed data and experimental splits are available from the corresponding author on reasonable request.

\bmhead{Materials availability}
All materials used in this study (such as trained models and configuration files) are available from the corresponding author on reasonable request.

\bmhead{Code availability}
The code used to implement the proposed method and run the experiments is available from the corresponding author on reasonable request.

\bibliography{references}

\end{document}